\documentclass[10pt,a4paper]{article}

\usepackage[margin=2.4cm]{geometry}
\usepackage{amsmath,amssymb}
\usepackage{newtxtext,newtxmath}
\usepackage{microtype}
\usepackage{booktabs}
\usepackage{array}
\usepackage{enumitem}
\usepackage{graphicx}
\usepackage{tikz}
\usetikzlibrary{positioning,arrows.meta,shapes.misc}
\usepackage[hidelinks]{hyperref}
\usepackage{caption}
\newcommand{\ind}{\mathbf{1}}

\title{\bfseries Artificial Intelligence-Assisted Digital Inventory of Cultural Heritage \& Traditional Knowledge\\ 
\medskip \small Case for Indonesian Open Digital Library of Culture}

\author{Hokky Situngkir\thanks{email: hs@compsoc.bandungfe.net, Dept. Computational Sociology - Bandung Fe Institute.}}

\begin{document}
\maketitle

\begin{abstract}
\noindent
The Indonesian Digital Library of Culture (\emph{Perpustakaan Digital Budaya
Indonesia}, PDBI; \texttt{budaya-indonesia.org}) is a participatory platform
that has collected tens of thousands of entries on Nusantara cultural heritage
through public contribution since 2007. Manual contribution faces three
structural barriers: \emph{coverage} (knowledge is scattered across languages
and sites), \emph{integrity} (open sources mix authentic documentation with
noise), and \emph{completeness} (subjects are recorded but their data remain
shallow). This paper presents a methodological framework for autonomous,
AI-based harvesting of cultural knowledge from the open web, designed to
\emph{expand} corpus coverage while \emph{intensifying} per-entry data depth.
The methodology is organised as a five-stage economic funnel---focused
crawling, multilingual extraction and canonicalisation, vector encoding with
blocking, agentic decision-making, and idempotent publication---under the
principle of \emph{deterministic orchestration, agentic decisions}. Each stage
is formalised: funnel economics and optimal filter ordering; crawl-frontier
dynamics as a subcritical branching process that explains the necessity of
recurrent re-seeding; fact-level novelty via a containment measure; Bayesian
multi-source evidence fusion with elevated publication thresholds for sacred
categories; exactly-once effects via idempotent upserts and the transactional
outbox; sliding-window inference budgeting with a reservation protocol;
statistical quality auditing; and seed selection as submodular coverage
maximisation. The framework retains four high-value human roles---curator of
direction, escalation approver, quality auditor, and guardian of
meaning---while machine autonomy is raised in stages. Ethical, legal, and
cultural-sensitivity implications are discussed, including the architectural
guarantee that the machine never overwrites human contributions.

\end{abstract}

\section{Introduction}

The documentation of intangible cultural heritage is internationally
recognised as a prerequisite of safeguarding \cite{unesco2003}. In Indonesia,
one of its institutional embodiments is the \textbf{Indonesian Digital Library
of Culture} (\emph{Perpustakaan Digital Budaya Indonesia}, PDBI) at
\texttt{budaya-indonesia.org}---a computational platform for participatory
preservation of traditional culture initiated in late 2007 by a research
community and cultural volunteers, later amplified by the ``One Million
Cultural Data'' movement \cite{situngkir2008,situngkir2010,kusuma2022,
unescopdbi}. To date PDBI holds tens of thousands of entries in fifteen
official categories, from musical instruments and folklore to rituals, dances,
and traditional medicine \cite{kusuma2022,unescopdbi}.

The participatory-manual model of collection, meritorious as it is, faces
three structural barriers. First, the \emph{coverage barrier}: documented
knowledge of Nusantara culture is scattered across the web---ethnographic
journals, colonial archives, regional-government sites, community and diaspora
documentation---in many languages, far beyond what human contributors can
sweep. Second, the \emph{integrity barrier}: open sources mix authentic
documentation with commercial content, ephemeral news, and unreferenced
claims; copying without filtering damages the corpus. Third, the
\emph{completeness barrier}: many entries are shallow---the subject is
recorded, but its recipe, motifs, performance structure, or regional
distribution are not.

This paper proposes and formalises a methodological framework that addresses
all three barriers through \emph{autonomous knowledge harvesting}: a system
that continuously discovers cultural knowledge artefacts on the open web in
any language, distils them into source-bound canonical records in Indonesian,
tests their novelty at the level of \emph{data} (not merely subjects), and
publishes only what genuinely adds---with architectural guarantees of quality,
traceability, and the sovereignty of human contribution. The framework has
been implemented in a production prototype operating against the PDBI corpus;
here the methodology is presented generically so that other digital
cultural-heritage documentation initiatives may replicate it.

The contributions are: (i) a five-stage \emph{economic funnel} architecture
that renders national-scale harvesting affordable
(\S\ref{sec:framework}, \S\ref{sec:funnel}); (ii) a complete mathematical
formalisation of every stage---from frontier dynamics as a branching process
(\S\ref{sec:crawl}) to Bayesian evidence fusion with graded thresholds for
sacred categories (\S\ref{sec:fusion}); (iii) exactly-once reliability
semantics and inference-budget control that guarantee operational
sustainability (\S\ref{sec:pub}--\S\ref{sec:budget}); and (iv) a staged
autonomy governance protocol with statistical auditing and well-defined human
roles (\S\ref{sec:audit}, \S\ref{sec:ops}).

\section{The Indonesian Digital Library of Culture as Context}
\label{sec:pdbi}

PDBI is built on the idea that the cultural diversity of the Indonesian
archipelago demands a participatory, web-based documentation system
\cite{situngkir2008,situngkir2010}. Kusumaningtiyas and Nurazizah
\cite{kusuma2022} document the role of society and communities---most notably
the Sobat Budaya community---in protecting and preserving Indonesian culture
through the platform, and its policy trajectory is recorded on UNESCO's policy
monitoring platform \cite{unescopdbi}. For the framework proposed here, PDBI
plays three methodological roles at once: (a) a \emph{target schema}---its
fifteen official categories and entry structure define the mapping space;
(b) a \emph{deduplication baseline}---the human-contributed corpus anchors the
question ``is this data already held?''; and (c) a \emph{publication
channel}---machine-produced entries are published back into PDBI through its
programmatic interface, making the machine an additional contributor alongside
(never a replacement for) human contributors.

Against taxonomies of digital-heritage platforms \cite{permatasari2020}, the
approach can be read as an evolutionary continuation from participatory
(web~2.0) platforms toward machine-intelligence-augmented platforms, without
abandoning the participatory principle: human labels and reviews become the
machine's training material (\S\ref{sec:audit}), in line with established
practice in cultural-heritage crowdsourcing \cite{ridge2014}.

\section{Related Work}
\label{sec:related}

\paragraph{Focused crawling.} Topic-focused web crawling with
relevance-prioritised frontiers was introduced by Chakrabarti et
al.~\cite{chakrabarti1999}; URL ordering by expected page value was studied by
Cho et al.~\cite{cho1998}, and large-scale crawler architecture with per-host
politeness by Heydon and Najork \cite{heydon1999}. The present framework
adopts a relevance-priority frontier with two domain-specific deviations: no
freshness re-crawling (cultural data are static), and the use of open
encyclopaedias strictly as \emph{pointers} to primary sources, never as
quotable sources.

\paragraph{Knowledge-base construction.} The never-ending language learner
NELL \cite{carlson2010} and Knowledge Vault \cite{dong2014} pioneered
web-scale fact extraction with probabilistic evidence fusion; this framework
continues that line with multilingual large language models
\cite{johnson2017,brown2020} as the extraction engine, adding a
\emph{source-boundness} constraint: facts without references are rejected by
the data schema itself.

\paragraph{Entity resolution and deduplication.} The classical theory of
record linkage is due to Fellegi and Sunter \cite{fellegi1969}; practical data
matching and blocking techniques are surveyed in
\cite{christen2012,papadakis2020}. The document-containment measure
\cite{broder1997} is adapted here to the level of fact sets. Approximate
nearest-neighbour search uses hierarchical navigable small-world graphs
(HNSW) \cite{malkov2020}.

\paragraph{Distributed-systems reliability.} The circuit-breaker pattern and
operational stability are treated by Nygard \cite{nygard2018}; idempotence as
the foundation of exactly-once effect semantics by Helland \cite{helland2012};
the transactional-outbox pattern and at-least-once delivery in
\cite{richardson2018,kleppmann2017}. Pipelined filter ordering follows
\cite{babu2004}; queue dimensioning uses Little's law \cite{little1961}.

\paragraph{Quality and calibration.} Error-proportion estimation with Wilson
intervals \cite{wilson1927}; confidence calibration of modern models
\cite{guo2017}; submodular maximisation for coverage selection
\cite{nemhauser1978}.

\section{Methodological Framework}
\label{sec:framework}

The framework's principal design rule can be stated in one sentence:
\emph{deterministic orchestration, agentic decisions}. The workflow is fixed,
auditable code---queues, transactions, retries, and logging are all
deterministic---while artificial-intelligence components (agents) are invoked
only at decision nodes that genuinely require reasoning: cultural relevance,
extraction--canonicalisation, novelty, and cross-validation. This separation
makes system behaviour predictable and inspectable, and concentrates inference
cost at high-value points only.

The architecture is a \textbf{five-stage funnel}
(Figure~\ref{fig:funnel}; Table~\ref{tab:stages}; notation in
Appendix~\ref{app:notation}). Early stages cost almost nothing per item and
filter aggressively; late stages are expensive but receive only survivors.
Stages are connected by durable message queues with one consumer per stage;
all shared state lives in a relational database with a vector extension, so
workers are fully stateless.

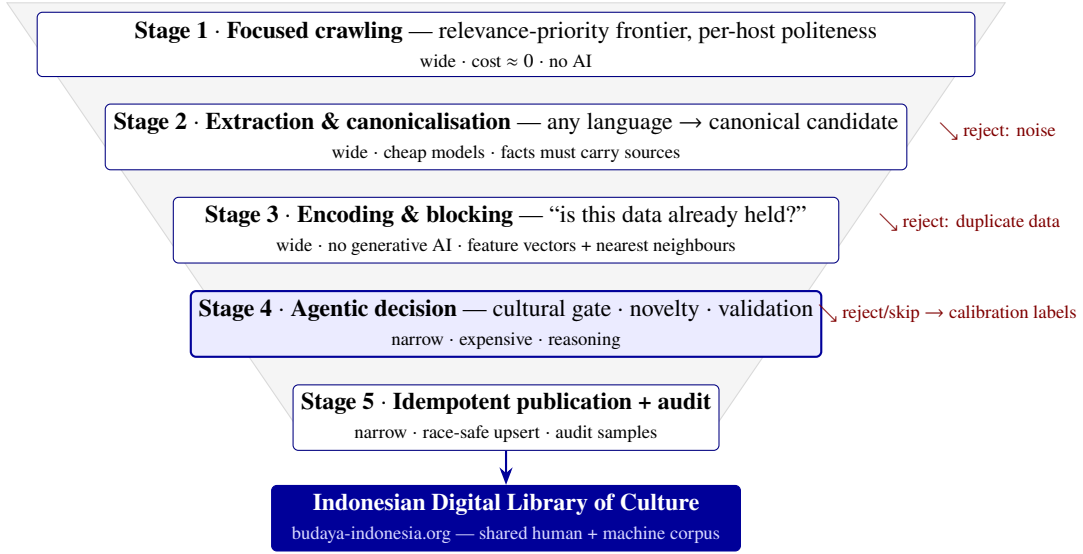
\begin{figure}[t]
\centering
\begin{tikzpicture}[
  font=\small,
  band/.style={draw=blue!50!black, fill=white, rounded corners=2pt,
               minimum height=8.5mm, align=center, inner sep=3pt},
  lbl/.style={font=\scriptsize, text=black!60},
  rej/.style={font=\scriptsize, text=red!50!black, anchor=west}
]
  \fill[black!4] (-6.6,0.55) -- (6.6,0.55) -- (2.6,-5.3) -- (-2.6,-5.3) -- cycle;
  \draw[black!15] (-6.6,0.55) -- (6.6,0.55) -- (2.6,-5.3) -- (-2.6,-5.3) -- cycle;
  \node[lbl, anchor=west, text=orange!60!black] at (-6.6,1.0)
    {\textbf{OPEN WEB} --- human-curated seeds; many languages};
  \node[band, minimum width=124mm] (s1) at (0,0)
    {\textbf{Stage 1 $\cdot$ Focused crawling} --- relevance-priority frontier, per-host politeness\\
     \scriptsize wide $\cdot$ cost $\approx 0$ $\cdot$ no AI};
  \node[band, minimum width=106mm, below=3.5mm of s1] (s2)
    {\textbf{Stage 2 $\cdot$ Extraction \& canonicalisation} --- any language $\to$ canonical candidate\\
     \scriptsize wide $\cdot$ cheap models $\cdot$ facts must carry sources};
  \node[band, minimum width=88mm, below=3.5mm of s2] (s3)
    {\textbf{Stage 3 $\cdot$ Encoding \& blocking} --- ``is this data already held?''\\
     \scriptsize wide $\cdot$ no generative AI $\cdot$ feature vectors + nearest neighbours};
  \node[band, minimum width=70mm, below=3.5mm of s3, fill=blue!8,
        draw=blue!60!black, thick] (s4)
    {\textbf{Stage 4 $\cdot$ Agentic decision} --- cultural gate $\cdot$ novelty $\cdot$ validation\\
     \scriptsize narrow $\cdot$ expensive $\cdot$ reasoning};
  \node[band, minimum width=54mm, below=3.5mm of s4] (s5)
    {\textbf{Stage 5 $\cdot$ Idempotent publication + audit}\\
     \scriptsize narrow $\cdot$ race-safe upsert $\cdot$ audit samples};
  \node[band, minimum width=62mm, below=4.5mm of s5, fill=blue!60!black,
        text=white, draw=blue!60!black] (pdbi)
    {\textbf{Indonesian Digital Library of Culture}\\
     \scriptsize\color{blue!15} budaya-indonesia.org --- shared human + machine corpus};
  \draw[-{Stealth}, blue!60!black, thick] (s5) -- (pdbi);
  \node[rej] at (5.6,-1.15) {$\searrow$ reject: noise};
  \node[rej] at (4.8,-2.35) {$\searrow$ reject: duplicate data};
  \node[rej] at (4.0,-3.55) {$\searrow$ reject/skip $\to$ calibration labels};
\end{tikzpicture}
\caption{The five-stage methodological funnel. Each stage narrows; expensive
reasoning only ever touches candidates that survived the cheap filters above
it. Rejections become calibration labels for subsequent learning.}
\label{fig:funnel}
\end{figure}

\begin{table}[t]
\centering\small
\begin{tabular}{@{}p{2.5cm}p{6.2cm}p{2.6cm}p{3.2cm}@{}}
\toprule
\textbf{Stage} & \textbf{Function} & \textbf{Cost profile} & \textbf{Core technology}\\
\midrule
1. Crawling & Discover pages, harvest links, filter early relevance & wide, $\approx$ zero/item & priority frontier, politeness\\
2. Extraction & Page $\to$ canonical Indonesian candidate, source-bound facts & wide, cheap models & multilingual LLM\\
3. Encoding & Feature vectors, blocking, neighbour search; structured evidence & wide, no LLM & partitioned ANN\\
4. Decision & Cultural gate, data-level novelty, cross-validation & narrow, expensive & reasoning agents\\
5. Publication & Structured record composition, licensed media, audit & narrow & idempotent upsert\\
\bottomrule
\end{tabular}
\caption{Summary of funnel stages and their cost profiles.}
\label{tab:stages}
\end{table}

\section{Formal Models}
\label{sec:models}

\subsection{Corpus formalisation and objective}
\label{sec:corpus}

The corpus is modelled as a set of entries $K=\{e\}$. Each entry is a tuple
\begin{equation}
e = (\iota_e, k_e, r_e, F_e), \qquad
f = (a, v, \sigma) \in F_e, \qquad \sigma(f) \neq \varnothing,
\label{eq:entry}
\end{equation}
where $\iota$ is the entry identity, $k \in \mathcal{K}$ the category
($|\mathcal{K}|=15$ for PDBI), $r \in \mathcal{R}$ the region, and $F$ a set
of facts with attribute $a$, value $v$, and source $\sigma$. The constraint
$\sigma \neq \varnothing$ (\emph{source-boundness}) is enforced by the data
schema: an unreferenced fact cannot be represented. The open web is modelled
as a directed graph $G=(V,E)$ of pages and links. The system objective is to
maximise validated information gain:
\begin{equation}
\max_{\pi}\;
\mathbb{E}\!\left[\sum_{e \in K'} w(e)\,\Delta I(e)\right]
\quad \text{s.t.} \quad C(\pi) \le B, \qquad \mathrm{Prec}(K') \ge p_{\min},
\label{eq:objective}
\end{equation}
where $\pi$ is the harvesting policy, $K'$ the published entries, $\Delta I$
the validated novel facts added, $w$ curatorial priority weights, $B$ the
budget, and $p_{\min}$ the minimum precision required by governance.

\subsection{Funnel economics and filter ordering}
\label{sec:funnel}

Let the funnel consist of $n$ stages with per-item costs $c_1,\dots,c_n$ and
pass rates $p_1,\dots,p_n$. The expected cost per input item and per published
entry are
\begin{equation}
C_{\mathrm{in}} = \sum_{k=1}^{n} c_k \prod_{j<k} p_j,
\qquad
C_{\mathrm{pub}} = C_{\mathrm{in}} \Big/ \prod_{k=1}^{n} p_k .
\label{eq:funnelcost}
\end{equation}
Because $\prod_{j<k} p_j$ shrinks rapidly, expensive stages (large $c_k$) must
sit behind strong filters. For filters whose order is exchangeable, the
ordering that minimises \eqref{eq:funnelcost} is ascending in the
cost--selectivity ratio \cite{babu2004}:
\begin{equation}
\text{order filters by ascending } \frac{c_i}{1-p_i}.
\label{eq:filterorder}
\end{equation}
This is the quantitative basis for placing a cheap lexical relevance gate at
Stage~1, AI-free deduplication at Stage~3, and expensive reasoning only at
Stage~4: the funnel drives $C_{\mathrm{pub}}$ down to a level that makes
national-scale harvesting feasible on a small, fixed daily inference budget
(\S\ref{sec:budget}).

\subsection{Crawl dynamics: priority, politeness, and frontier extinction}
\label{sec:crawl}

The frontier $\Phi$ is a priority queue of normalised URLs (unique key;
idempotent under rediscovery). Each URL's priority is a function of relevance
features:
\begin{equation}
U(u) = \sum_{m} w_m\, x_m(u),
\label{eq:priority}
\end{equation}
where the features $x_m$ include link-context score, page type (index/listing
pages are treated as \emph{discovery-only}: their links are harvested, the
pages themselves never become entries), depth from seed (bounded by
$d \le d_{\max}$), and denylist membership. Crawl politeness is enforced as a
minimum per-host delay $\delta$, bounding per-host and global fetch rates:
\begin{equation}
\lambda_h \le 1/\delta \;\;\forall h,
\qquad
\Lambda \le |H_{\mathrm{active}}| / \delta,
\label{eq:politeness}
\end{equation}
making $\delta$ an ethical-legal constraint and a capacity parameter at once
\cite{heydon1999}. Two domain-specific decisions distinguish cultural from
news crawling: (i) \emph{no freshness re-crawling}---cultural data are static,
old pages are never re-fetched; (ii) \emph{open encyclopaedias as
doorways}---the references and external-links sections of encyclopaedia
articles are harvested and followed to primary sources, while encyclopaedia
text itself is never quoted or stored.

The structural consequence of these decisions can be modelled as a
Galton--Watson branching process \cite{harris1963}. Let $Z_t$ be the number of
productive URLs in generation $t$, and $\xi$ the number of new, unique,
filter-passing links yielded by one page:
\begin{equation}
Z_{t+1} = \sum_{i=1}^{Z_t} \xi_i,
\qquad
m = \mathbb{E}[\xi].
\label{eq:branching}
\end{equation}
Since URL deduplication and the saturation of seed neighbourhoods drive $m$
down over time, the process becomes subcritical ($m<1$) and dies out almost
surely: the extinction probability $q$ is the smallest fixed point of the
offspring probability generating function $g$,
\begin{equation}
q = g(q), \qquad q = 1 \iff m \le 1 .
\label{eq:extinction}
\end{equation}
\emph{Frontier exhaustion is therefore not an incidental failure but a
structural property} of any system without freshness re-crawling. The protocol
consequently makes re-seeding an operational cadence: new seeds are injected
(per category ``tranche'', one tranche at a time) whenever frontier depth
falls below a threshold,
\begin{equation}
|\Phi_t| < \phi_{\min} \;\Rightarrow\; \text{inject next seed tranche},
\label{eq:reseed}
\end{equation}
while monitoring inter-tranche yield before the next widening
(\S\ref{sec:coverage} formalises tranche selection).

\subsection{Multilingual extraction and canonicalisation}
\label{sec:extract}

Stage~2 maps a raw document in any language to zero or more canonical
candidates:
\begin{equation}
\psi : d \longmapsto \{x_1,\dots,x_j\},
\qquad
x = (n, A, k, r, F),
\qquad
T : \Sigma^{*}_{\mathrm{any}} \to \Sigma^{*}_{\mathrm{id}},
\label{eq:extraction}
\end{equation}
where $n$ is the canonical entity name with aliases $A$, $k$ the category, $r$
the region, and $F$ facts all satisfying the source-boundness constraint
\eqref{eq:entry}. The language-normalisation operator $T$ (implemented by
multilingual language models \cite{johnson2017,brown2020}) is placed
\emph{once, in one place}: all downstream stages operate on a single canonical
language, so matching, reasoning, and publication are language-clean. Long
documents are segmented before extraction (one source $\to$ many candidates)
with checkpoints so processing is monotone and resumable. Media (images) are
captured as \emph{references} (URL, caption, licence hint); binary retrieval
is deferred until a candidate is accepted (fetch-on-accept), reducing cost and
licensing risk.

\subsection{Vector representation, blocking, and partitioned neighbour search}
\label{sec:encode}

Every candidate and entry is represented by an explicit two-block feature
vector:
\begin{equation}
\varphi(x) = [\varphi_{\mathrm{shape}}(x);\, \varphi_{\mathrm{content}}(x)],
\label{eq:features}
\end{equation}
with a \emph{shape} block (category, region, structural flags) and a
\emph{content} block (character $n$-grams of the entity name, region tokens,
category-conditional attribute bags). Candidate pairs are compared only if
they pass the blocking predicate \cite{christen2012,papadakis2020}:
\begin{equation}
B(x,y) = \ind\!\left[\,k_x = k_y \,\wedge\, r_x \sim r_y\,\right],
\label{eq:blocking}
\end{equation}
and the final similarity is computed \emph{from the content block only}---a
convex combination of $n$-gram cosine similarity, attribute Jaccard, and name
match:
\begin{equation}
s(x,y) = \alpha \cos\!\big(\varphi_{\mathrm{c}}(x), \varphi_{\mathrm{c}}(y)\big)
       + \beta\, J(A_x, A_y)
       + \gamma\, \mathrm{sim}_{\mathrm{name}}(n_x, n_y),
\qquad \alpha+\beta+\gamma = 1 .
\label{eq:similarity}
\end{equation}
The shape block is deliberately never fused into the score: its cardinality is
low, so its discriminative power decays as the corpus grows---in a corpus of
size $N$ with $P=|\mathcal{K}|\cdot|\mathcal{R}|$ balanced partitions, the
number of same-category-same-region pairs grows quadratically,
$\mathbb{E}[\text{pairs/partition}] \approx N^2/(2P^2)$, so
``same category and region'' carries almost no information at large $N$.
Conversely the shape key is highly effective as a \emph{partitioner}: the
nearest-neighbour index is partitioned by $(k,r)$---the same key serving
blocking accuracy and sharding strategy---with one HNSW graph per partition
\cite{malkov2020} and search complexity
\begin{equation}
O(\log N_p), \qquad N_p \approx N/P .
\label{eq:hnsw}
\end{equation}
The stage's output is not an opaque scalar score but \emph{structured
evidence} per neighbour (category-region agreement, attribute overlap, name
match) that the next stage can reason over and humans can audit.

\subsection{Data-level novelty: fact containment}
\label{sec:novelty}

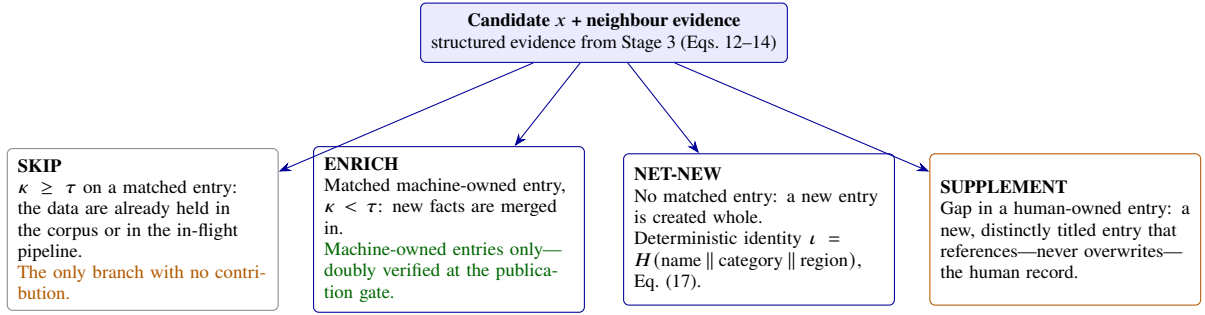
\begin{figure}[t]
\centering
\begin{tikzpicture}[
  font=\scriptsize,
  root/.style={draw=blue!60!black, fill=blue!8, rounded corners=2pt,
               align=center, inner sep=4pt},
  leaf/.style={draw, rounded corners=2pt, fill=white, align=left,
               inner sep=4pt, text width=33mm, minimum height=20mm},
  arr/.style={-{Stealth}, blue!60!black}
]
  \node[root] (r) at (0,0)
    {\textbf{Candidate $x$ + neighbour evidence}\\
     structured evidence from Stage 3 (Eqs.~\ref{eq:blocking}--\ref{eq:hnsw})};
  \node[leaf, draw=black!45] (skip) at (-6.1,-2.6)
    {\textbf{SKIP}\\ $\kappa \ge \tau$ on a matched entry: the data are
     already held in the corpus or in the in-flight pipeline.\\
     {\color{orange!70!black}The only branch with no contribution.}};
  \node[leaf, draw=blue!60!black] (enr) at (-2.05,-2.6)
    {\textbf{ENRICH}\\ Matched machine-owned entry, $\kappa<\tau$: new facts
     are merged in.\\
     {\color{green!40!black}Machine-owned entries only---doubly verified at
     the publication gate.}};
  \node[leaf, draw=blue!60!black] (new) at (2.05,-2.6)
    {\textbf{NET-NEW}\\ No matched entry: a new entry is created whole.\\
     Deterministic identity $\iota = H(\text{name}\,\|\,\text{category}\,\|\,
     \text{region})$, Eq.~\eqref{eq:identity}.};
  \node[leaf, draw=orange!70!black] (sup) at (6.1,-2.6)
    {\textbf{SUPPLEMENT}\\ Gap in a human-owned entry: a new, distinctly
     titled entry that references---never overwrites---the human record.};
  \draw[arr] (r) -- (skip);
  \draw[arr] (r) -- (enr);
  \draw[arr] (r) -- (new);
  \draw[arr] (r) -- (sup);
\end{tikzpicture}
\caption{The novelty decision tree, Eq.~\eqref{eq:novelty}. Expansion is
realised by the \emph{net-new} branch; intensification by the \emph{enrich}
and \emph{supplement} branches. Three of four branches contribute: data-level
duplication turns an anti-duplicate engine into an intensification engine.}
\label{fig:novelty}
\end{figure}

The framework's most important methodological differentiator is its definition
of duplication. Conventional deduplication asks ``does this subject already
exist?''; this framework asks \emph{``does the corpus already hold this
data?''} Adapting the document-containment measure \cite{broder1997} to fact
sets, for candidate $x$ and linked entry $e$ define the fact containment
\begin{equation}
\kappa(x,e) = \frac{\lvert F_x \sqcap F_e \rvert}{\lvert F_x \rvert},
\label{eq:containment}
\end{equation}
where $\sqcap$ intersects facts after canonical attribute alignment. The
novelty decision tree (Figure~\ref{fig:novelty}) is
\begin{equation}
\text{decide}(x) =
\begin{cases}
\textit{skip}, & \exists\, e \text{ matched with } \kappa(x,e) \ge \tau;\\[2pt]
\textit{enrich}, & \exists\, e \text{ matched, machine-owned, } \kappa(x,e) < \tau;\\[2pt]
\textit{supplement}, & \exists\, e \text{ matched, human-owned, } \kappa(x,e) < \tau;\\[2pt]
\textit{net-new}, & \nexists\, e \text{ matched}.
\end{cases}
\label{eq:novelty}
\end{equation}
The \emph{supplement} rule enforces the sovereignty of human contribution
architecturally: update operations may only touch machine-owned entries
(doubly verified at the publication gate); a genuine gap in a human-covered
subject is filled by a new, distinctly titled entry that references---never
overwrites---the human record. Entry identity is deterministic in content,
\begin{equation}
\iota = H\big(\mathrm{norm}(n) \,\|\, k \,\|\, r\big),
\label{eq:identity}
\end{equation}
so many sources about the same variant accumulate into one entry, while
distinct regional variants---precisely the local diversity that preservation
seeks to protect---receive entries of their own.

\subsection{Multi-source evidence fusion and graded publication thresholds}
\label{sec:fusion}

Before publication, every fact is cross-validated against its clustered
sources. Under conditional independence of sources, the probability that fact
$f$ is correct given sources $S=\{s_1,\dots,s_k\}$ takes odds form
\cite{dong2014,fellegi1969}:
\begin{equation}
O(f \mid S) = O_0 \prod_{i=1}^{k} \lambda_i,
\qquad
\lambda_i =
\frac{P(s_i \mid f \text{ true})}{P(s_i \mid f \text{ false})},
\label{eq:odds}
\end{equation}
where $\lambda_i$ is a source-credibility factor (a function of source type:
academic, official institution, community). A single credible, uncontradicted
source may be accepted; contradiction lowers the odds. The publication rule is
graded by category:
\begin{equation}
\text{auto-publish} \iff P(f \mid S) \ge \theta_k;
\qquad
\text{sensitive: } \theta_s > \theta \,\wedge\, |S_{\mathrm{indep}}| \ge 2,
\;\text{else} \to \text{human review},
\label{eq:publish}
\end{equation}
with sensitive categories (rituals; sacred objects and dances) designated
together with cultural stakeholders, not by the machine. Low-confidence
decisions that are still published are flagged and prioritised for audit
sampling (\S\ref{sec:audit})---transparency of uncertainty replaces blind
trust.

\subsection{Idempotent publication and exactly-once effect semantics}
\label{sec:pub}

Distributed messaging can guarantee only \emph{at-least-once} delivery;
\emph{exactly-once effects} are obtained by making publication idempotent
\cite{helland2012,kleppmann2017}. With the upsert operator $U$ keyed by
identity \eqref{eq:identity}:
\begin{equation}
U\big(U(K,x),\,x\big) = U(K,x),
\label{eq:idempotent}
\end{equation}
redelivery leaves the final state unchanged. Consistency between database and
message queue is guaranteed by the \emph{transactional outbox} pattern
\cite{richardson2018}: the stage artefact, the follow-on job, and the outgoing
message are written in \emph{one} transaction; a relay moves messages to the
queue and marks them sent only after the queue acknowledges. Transient
failures are handled by bounded geometric backoff:
\begin{equation}
d_n = \min\!\big(d_0\gamma^{\,n},\, d_{\max}\big),
\quad n \le M;
\qquad
n > M \Rightarrow \text{dead-letter queue},
\label{eq:backoff}
\end{equation}
with a dead-letter queue that can be re-driven after capability
recovery---work is never lost, only delayed.

\subsection{Inference-cost control and provider reliability}
\label{sec:budget}

\begin{figure}[t]
\centering
\begin{tikzpicture}[
  font=\scriptsize,
  st/.style={draw, rounded rectangle, minimum width=30mm, minimum height=10mm,
             align=center, inner sep=3pt},
  arr/.style={-{Stealth}, blue!60!black}
]
  \node[st, draw=green!45!black] (c) at (0,0)
    {\textbf{\color{green!45!black}CLOSED}\\ normal traffic};
  \node[st, draw=red!55!black, right=22mm of c] (o)
    {\textbf{\color{red!55!black}OPEN}\\ fail fast; work buffered};
  \node[st, draw=orange!70!black, right=22mm of o] (h)
    {\textbf{\color{orange!70!black}HALF-OPEN}\\ one trial call};
  \draw[arr] (c) -- node[above]{failures $\ge$ threshold} (o);
  \draw[arr] (o) -- node[above]{cooldown $\Delta$} (h);
  \draw[arr] (h.south) .. controls +(0,-1.2) and +(0,-1.2) ..
    node[below]{successful probe $\to$ close} (c.south);
  \draw[arr] (h.north) .. controls +(-0.8,0.8) and +(0.8,0.8) ..
    node[above]{probe fails} (o.north);
\end{tikzpicture}
\caption{The per-provider circuit-breaker state machine \cite{nygard2018}.
Environmental failure leads to buffer-and-retry, never to discarding work;
error classes are distinguished (rate-limit $\to$ backoff; budget $\to$ stage
pause; moderation $\to$ per-request terminal).}
\label{fig:breaker}
\end{figure}
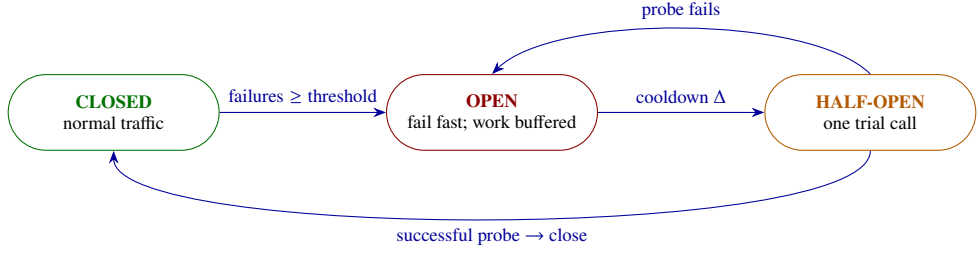

All external capabilities (page fetching, search, AI inference, translation)
are wrapped in one uniform provider layer with health monitoring, circuit
breakers, and failover \cite{nygard2018}. The circuit breaker is modelled as a
three-state machine (Figure~\ref{fig:breaker}): \emph{closed} $\to$
\emph{open} after consecutive failures exceed a threshold; \emph{open} $\to$
\emph{half-open} after a cooldown $\Delta$; one successful probe closes it
again. The inference budget is enforced as a sliding-window constraint with a
reservation protocol:
\begin{equation}
\sum_{t' \in (t-W,\,t]} \mathrm{spend}(t') \;+\; \mathrm{reserved}_{\mathrm{in\text{-}flight}}
\;\le\; B_W,
\label{eq:budget}
\end{equation}
with a shared cost ledger as the single source of truth: every worker
\emph{reserves before calling} and \emph{settles afterwards}, so $K$ parallel
workers enforce one cap, not $K$ caps. The sliding-window form (rather than a
lifetime accumulation) makes the constraint self-clearing as old spend leaves
the window---the system cannot freeze permanently on account of its history.
Per-stage concurrency is dimensioned by Little's law \cite{little1961},
$L = \lambda \bar{W}$, with bounded per-stage concurrency as the backpressure
mechanism.

\subsection{Statistical auditing, calibration, and feedback}
\label{sec:audit}

Every publication emits an audit sample into a human review queue, with
inclusion probability boosted for low-confidence decisions and sensitive
categories (stratified sampling). The corpus error proportion $\hat{p}$ is
estimated with Wilson confidence intervals \cite{wilson1927}; the sample size
for margin $\varepsilon$ at level $z$ is
\begin{equation}
n \ge z^2\, p(1-p) / \varepsilon^2 .
\label{eq:samplesize}
\end{equation}
Model confidence quality is monitored with the expected calibration error
\cite{guo2017}:
\begin{equation}
\mathrm{ECE} = \sum_b \frac{|B_b|}{n}\,
\big|\, \mathrm{acc}(B_b) - \mathrm{conf}(B_b) \,\big| .
\label{eq:ece}
\end{equation}
Human review labels (correct/incorrect; duplicate/novel) are written back to
the decision store and become training data for the next-generation learned
similarity encoder---a layer that is \emph{earned} from operation, not built
up front. A second feedback loop is structural: every published entry is
immediately re-encoded into the index (\S\ref{sec:encode}), so the standard of
``what is already held'' rises as the corpus grows.

\subsection{Coverage and submodular seed selection}
\label{sec:coverage}

Corpus coverage is mapped by a matrix $M[k,r]$ over category $\times$ region.
The coverage value of a seed set $S$ is defined as
\begin{equation}
f(S) = \sum_{(k,r)} w_{k,r}\,
\min\!\left\{1,\; n_{k,r}(S)/n^{*}_{k,r}\right\},
\label{eq:coverage}
\end{equation}
where $n_{k,r}(S)$ is the yield of cell $(k,r)$ from seeds $S$ and $n^{*}$ a
per-cell target. Such a sum-of-minima function is monotone and submodular
(diminishing returns), so greedy tranche selection---adding the tranche with
the largest marginal yield per unit cost, one tranche at a time while
observing outcomes---guarantees a $(1-1/e)$ approximation of the optimum
\cite{nemhauser1978}:
\begin{equation}
f(S_{\mathrm{greedy}}) \ge \Big(1 - \tfrac{1}{e}\Big)\,
\max_{|S| \le B} f(S).
\label{eq:greedy}
\end{equation}
The operational protocol of ``widening one tranche at a time under yield
monitoring'' is therefore not mere caution but an implementation of the greedy
algorithm on a submodular coverage function---and simultaneously the feedback
mechanism for estimating true marginal yields.

\section{Operational Protocol and Autonomy Governance}
\label{sec:ops}

The methodology runs as a recurring seven-step protocol:
\begin{enumerate}[leftmargin=1.6em, itemsep=1pt]
\item \textbf{Seed curation.} Humans compose and widen the seed list per
  category tranche (\S\ref{sec:coverage}), guided by the coverage matrix
  $M[k,r]$.
\item \textbf{Focused crawling} with the priority frontier
  \eqref{eq:priority}, politeness \eqref{eq:politeness}, depth and page-budget
  bounds; frontier depth is monitored against the re-seeding threshold
  \eqref{eq:reseed}.
\item \textbf{Extraction--canonicalisation} across languages
  \eqref{eq:extraction} under the source-boundness constraint
  \eqref{eq:entry}.
\item \textbf{Encoding and filtering}---blocking \eqref{eq:blocking}, content
  similarity \eqref{eq:similarity}, partitioned neighbour search
  \eqref{eq:hnsw}, structured evidence.
\item \textbf{Agentic decision}---the cultural scope gate (a specific,
  \emph{named} item of Nusantara tradition, mappable to a category; with an
  explicit carve-out for diaspora heritage practised today as Indonesian
  culture), the novelty tree \eqref{eq:novelty}, evidence fusion and graded
  thresholds \eqref{eq:odds}--\eqref{eq:publish}.
\item \textbf{Idempotent publication} \eqref{eq:idempotent} with the ownership
  gate, licensed media retrieval, and audit-sample emission.
\item \textbf{Audit and learning}---error estimation \eqref{eq:samplesize},
  calibration monitoring \eqref{eq:ece}, label write-back.
\end{enumerate}

\paragraph{Staged autonomy.} The machine is raised through three
modes---\emph{shadow} (compose and validate, no writes), \emph{trickle}
(sampled $N$/day publications under observation), and \emph{full}
(autonomous)---with human sign-off at each promotion, gated by three risk
classes: cultural correctness and sensitivity; the safety of the shared corpus
(a dedicated machine account, add-never-overwrite, reconcile-before-write);
and legal, terms-of-service, and licensing compliance. In steady state no
human gate blocks the daily flow, yet four human roles remain decisive: the
\emph{curator of direction} (seeds and tranches), the \emph{escalation
approver} (autonomy modes), the \emph{quality auditor} (samples and labels),
and the \emph{guardian of meaning} (scope definition, the sensitive-category
list, the diaspora carve-out).

\section{Discussion}
\label{sec:discussion}

\paragraph{Expansion and intensification as two outputs of one machine.}
In this framework, \emph{expansion} (new entries; widening cells of the
coverage matrix) and \emph{intensification} (deepening facts of existing
entries via the \emph{enrich}/\emph{supplement} branches) are not two separate
programmes but two branches of the same decision tree \eqref{eq:novelty}---
both flowing from the data-level definition of duplication
\eqref{eq:containment}. This directly answers the coverage and completeness
barriers at once.

\paragraph{Integrity as an architectural property.} Source-boundness is
enforced by the schema \eqref{eq:entry}; correctness is treated
probabilistically with explicit thresholds
\eqref{eq:odds}--\eqref{eq:publish}; uncertainty is flagged and audited
(\S\ref{sec:audit}); correction is always possible (idempotent remediation and
soft deletion); and human contributions are structurally untouchable
\eqref{eq:novelty}. Integrity is thus not a guideline hoping to be obeyed but
an invariant enforced by code.

\paragraph{Limitations.} First, frontier extinction \eqref{eq:extinction}
demands continuous seed curation; automating the trigger \eqref{eq:reseed}
reduces but does not remove the curator's role. Second, some sources behind
anti-bot protection are unreachable without technical escalation that is
costly and brittle---protection circumvention is an arms race with no finish
line, and its ethical-legal boundary must be set by policy, not technique.
Third, the conditional-independence assumption in \eqref{eq:odds} is violated
when sources copy one another; estimating source kinship (copy clusters) is
future work. Fourth, cross-lingual canonicalisation quality depends on the
language models used and requires dedicated evaluation for regional languages
with non-Latin scripts. Fifth, the audit-label loop must actually be closed
for the learned encoder to be earned; deployment experience shows this step is
easily postponed and deserves treatment as a governance indicator.

\paragraph{Ethics and cultural sensitivity.} The framework translates
sensitivity into mechanism: higher thresholds and multi-source requirements
for sacred categories \eqref{eq:publish}, human review routes, and stakeholder
designation of the sensitive list. Crawl politeness \eqref{eq:politeness} and
media licence checking are treated as hard constraints. The general principle:
cultural values govern the technology, not the other way around.

\section{Conclusion}
\label{sec:conclusion}

This paper has presented a complete methodological framework---together with
its formal models---for expanding and intensifying digital cultural-heritage
documentation through autonomous, AI-based knowledge harvesting, with the
Indonesian Digital Library of Culture as the application context. Three ideas
form its backbone: the \emph{economic funnel} that reserves expensive
reasoning for filtered candidates
\eqref{eq:funnelcost}--\eqref{eq:filterorder}; \emph{data-level novelty} that
turns deduplication into an intensification engine
\eqref{eq:containment}--\eqref{eq:novelty}; and \emph{reliability and
governance as architectural invariants}---from exactly-once effects
\eqref{eq:idempotent} and sliding-window budgets \eqref{eq:budget} to graded
thresholds for sacred categories \eqref{eq:publish} and statistical audits
\eqref{eq:samplesize}--\eqref{eq:ece}. The framework demonstrates that
artificial intelligence can extend a cultural encyclopaedia's reach to scales
impossible for manual curation without displacing the sovereignty of human
contributors---indeed it elevates the human role to curator of direction,
approver, auditor, and guardian of meaning. Future work includes the learned
similarity encoder trained from audit labels, source-kinship estimation for
evidence fusion, and evaluation of regional-language canonicalisation.

\section*{Acknowledgements}

The author thanks colleagues in Svadaya Budhi Futura Incresca currently administering the Indonesian
Digital Library of Culture (\emph{Perpustakaan Digital Budaya Indonesia})---in
particular the Sobat Budaya community, whose participatory corpus provides both the target schema and the deduplication
baseline for the framework described here---and the cultural stakeholders
whose counsel shapes the sensitive-category list and the scope rules. The
sovereignty of human contribution that this framework protects is, first of
all, theirs.

\clearpage
\appendix
\section{Notation}
\label{app:notation}

\begin{table}[h]
\centering\small
\begin{tabular}{@{}p{3.4cm}p{11.4cm}@{}}
\toprule
\textbf{Symbol} & \textbf{Meaning}\\
\midrule
$e$, $x$; $F$, $f$ & corpus entry, candidate; fact set, fact (attribute, value, source)\\
$\iota$, $k$, $r$ & entry identity; category ($\in\mathcal{K}$, $|\mathcal{K}|=15$); region ($\in\mathcal{R}$)\\
$c_k$, $p_k$ & per-item cost and pass rate of funnel stage $k$\\
$\Phi$, $U(u)$, $\delta$, $d_{\max}$ & frontier; URL priority; per-host politeness delay; maximum depth\\
$m$, $q$, $g$ & offspring mean, extinction probability, generating function of the branching process\\
$\varphi$, $B(\cdot,\cdot)$, $s(\cdot,\cdot)$, $\kappa$ & feature vector; blocking predicate; content similarity; fact containment\\
$\lambda_i$, $\theta_k$, $\theta_s$ & source-credibility factor; category publication threshold; sensitive-category threshold\\
$U$, $d_n$, $M$ & idempotent upsert operator; backoff schedule; redelivery bound\\
$B_W$, $W$ & per-window budget cap; sliding-window width\\
$f(S)$, $M[k,r]$ & submodular coverage function over seeds; category $\times$ region coverage matrix\\
\bottomrule
\end{tabular}
\caption{Principal notation.}
\label{tab:notation}
\end{table}

\end{document}